\documentclass{article}
\usepackage{iclr2027_conference,times}
\usepackage[T1]{fontenc}
\usepackage[utf8]{inputenc}
\usepackage{amsmath,amssymb,graphicx,booktabs,tabularx,longtable,array}
\usepackage{url,xurl}
\usepackage[letterpaper,left=1.5in,right=1.5in,top=1in,bottom=1.05in,
includeheadfoot,headheight=12pt,headsep=18pt,footskip=24pt]{geometry}
\usepackage{hyperref}
\hypersetup{hidelinks,pdftitle={Natural State-Prediction Accuracy Can Hide Weak Controlled Responsiveness in VLA Readouts},pdfauthor={Hyungjoon Kim; Wonbin Son; Mi Young Lee; Jun Young Lee; Seungmin Rho},pdfsubject={Preprint}}
\newcolumntype{L}[1]{>{\raggedright\arraybackslash}p{#1}}
\newcolumntype{Y}{>{\raggedright\arraybackslash}X}

\title{Natural State-Prediction Accuracy Can Hide Weak Controlled Responsiveness in VLA Readouts}
\newcommand{\authorline}[1]{\makebox[\dimexpr\textwidth-2\tabcolsep\relax][c]{#1}}
\author{%
\authorline{\bfseries Hyungjoon Kim\textsuperscript{1}\thanks{First author.}\quad
Wonbin Son\textsuperscript{1}\quad Mi Young Lee\textsuperscript{2}}\\[3pt]
\authorline{\bfseries Jun Young Lee\textsuperscript{2}\quad
Seungmin Rho\textsuperscript{2}\thanks{Corresponding author: \href{mailto:smrho@cau.ac.kr}{\texttt{smrho@cau.ac.kr}}.}}\\[6pt]
\authorline{\normalfont\textsuperscript{1}\,Changwon National University\qquad
\textsuperscript{2}\,Chung-Ang University}\\[4pt]
\authorline{\normalfont\small\texttt{hyungjoon@changwon.ac.kr}\quad
\texttt{diwjidghk78@gmail.com}}\\
\authorline{\normalfont\small\texttt{miylee@cau.ac.kr}\quad
\texttt{tfg0074@cau.ac.kr}\quad\texttt{smrho@cau.ac.kr}}
}
\iclrfinalcopy
\begin{document}
\maketitle
\lhead{Preprint}
\begin{abstract}
Accurately decoding object states from the internal representations of vision-language-action (VLA) models does not establish that the predictions respond faithfully to changes in the target physical state. In natural observations, object state, robot configuration, occlusion, and task progress vary together, allowing contextual cues to contribute to prediction. In this paper, we introduce an evaluation framework that separates prediction accuracy, target-state responsiveness, and context stability using physically validated observations that cross target coordinates with robot contexts. We demonstrate that high natural-trajectory accuracy can coexist with weak controlled target-state responsiveness in fixed representation--readout pairs. Comparisons and interventions involving representations, readouts, and training data show that the three properties provide distinct diagnostic information. Furthermore, adding responsiveness and context sensitivity to a failure predictor based on initial state error and physical variables reduces policy-failure prediction error on new initializations relative to the specified baseline while same-observation controlled MAE is also informative. These findings motivate evaluating target-state responsiveness and context stability alongside natural prediction accuracy, and examining their relationship to actual policy behavior and task outcomes.
\end{abstract}

\section{Introduction}
Recent studies examine whether object states, robot states, and task progress can be decoded from vision-language-action (VLA) representations. They recover object and action states, state transitions, and robot-related variables from hidden features, and use these estimates for behavioral intervention or failure recovery~\citep{lu2025probing,molinari2025emergent,buurmeijer2026observing,zhang2026probeact}. Such analyses provide a way to investigate the state information available in a model and its potential use.

However, low prediction error on natural trajectories alone does not establish that a fixed representation--readout pair responds faithfully to the target physical variable. Along successful trajectories, object state changes together with robot configuration, contact, occlusion, and task progress. Natural accuracy therefore leaves unresolved whether predictions respond appropriately when object state changes at fixed robot configuration, or remain stable when robot configuration changes at fixed object state.

We distinguish three properties. \textbf{Prediction accuracy} measures how accurately a readout estimates the target coordinate in natural observations. \textbf{Target-state responsiveness} measures how its predictions change with the actual coordinate at fixed robot context. \textbf{Context stability} measures how consistently it predicts the same target state across robot contexts. Responsiveness and stability must be considered jointly: an almost constant readout can appear stable while failing to respond to state changes, whereas a responsive readout can remain sensitive to context. Together with natural accuracy, these properties assess how faithfully the readout responds to the target physical coordinate.

We fit state readouts to frozen VLA features and evaluate each pair on natural observations and observations that cross target coordinates with robot contexts. We also examine the relationship between these diagnostics and policy outcomes. Decoding a state, responding to a physical change, producing an action, and completing a task are distinct evaluation targets.

\begin{figure}[tb]
\centering
\includegraphics[width=\linewidth]{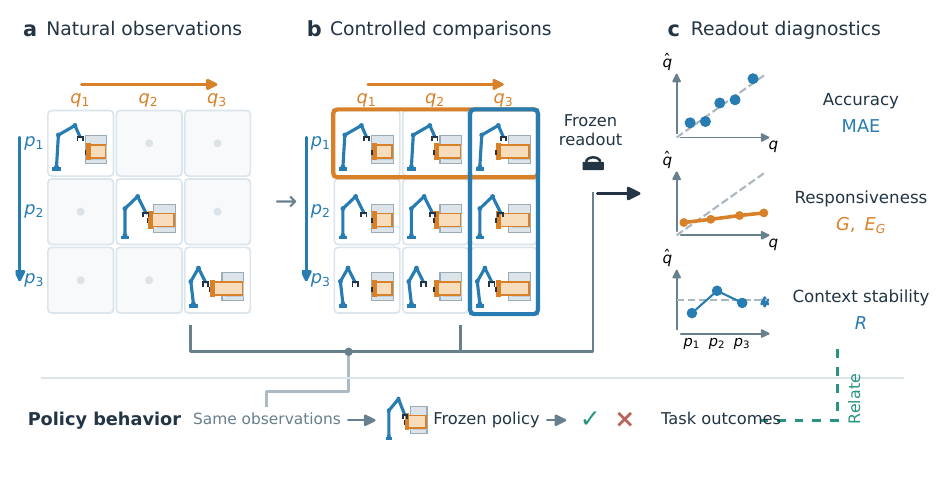}
\caption{Evaluation framework. (a) Target state and robot context co-vary in natural observations. (b) Crossed comparisons vary state at fixed context (orange row) and context at fixed state (blue column). (c) The same fixed representation--readout pair is evaluated for accuracy, responsiveness, and stability. A separate fixed policy receives observations, and its outcomes are related to the diagnostics (dashed connection). Readout predictions are not policy inputs. Scenes, grids, and small plots are schematic illustrations.}
\label{fig:framework}
\end{figure}

Figure~\ref{fig:framework} summarizes the framework and its behavioral connection. First, we construct physically and visually validated crossed comparisons that separate accuracy, responsiveness, and context stability. Second, a fixed confirmatory protocol and comparisons across representations and readouts establish the separation between natural accuracy and controlled responsiveness; training interventions and candidate comparisons reveal how the diagnostics change under different choices. Third, we evaluate failure predictors on separate initializations to determine whether controlled diagnostics provide outcome information beyond initial state error and specified physical variables. A comparison with simple error summaries from the same controlled observations further characterizes this contribution.

\section{Related Work}
\paragraph{Decoding state and task signals from VLA representations.}
\citet{lu2025probing} analyze symbolic object and action states in OpenVLA, while \citet{molinari2025emergent} probe future changes in visual state embeddings. \citet{buurmeijer2026observing} study linearly observable robot-related features and representation interventions for behavioral control. ProbeAct uses object-position estimates from hidden features for failure recovery~\citep{zhang2026probeact}. \citet{bhardwaj2026decoding} analyze task progress and language counterfactuals, distinguishing decodability from steerability. \citet{zhang2026frozen} evaluate success-related signals in frozen representations through task- and timestep-matched comparisons and action selection.

\paragraph{State representations and control performance.}
\citet{dong2026capturing} relate the quality of environment-state decoding from pretrained visual encoders to downstream control performance. This supports using state information when selecting representations. We address a different question from correlations across models: whether a given readout's natural accuracy establishes its responsiveness to controlled physical changes.

\paragraph{Interpreting and intervening on VLA computations.}
\citet{haon2025mechanistic} steer behavior through semantic activation directions, and \citet{mitra2025mechanistic} identify task-relevant attention heads for selective finetuning. \citet{swann2026sparse} study interpretable sparse-autoencoder features and their behavioral effects. \citet{grant2026features} examine modality and computational pathways through activation injection and probing. VLA-Trace connects representation changes, attention interventions, and behavioral tests~\citep{shi2026vlatrace}, while \citet{zhang2026embodied} study how masking visual regions affects actions and relates to generalization.

\paragraph{Diagnosing robot-policy failures.}
DART~\citep{laskey2017dart} and HYDRA~\citep{belkhale2023hydra} address execution errors and distribution shift in imitation learning. Sentinel monitors generative policies using consistency and progress~\citep{agia2025sentinel}. Complementing these studies, we measure the responsiveness and stability of the same state readout through crossed physical-coordinate and robot-context comparisons, and test whether these summaries provide outcome information beyond initial state error and physical variables.

\section{Method}
\label{sec:method}
The proposed evaluation separates natural state-prediction accuracy from responses to controlled target-coordinate changes. Let $z=f(o)$ be a frozen representation and $\hat q=h(z)$ a fitted state readout. We apply the same $h\circ f$ to natural and crossed target-state-by-context observations. Responsiveness is measured at the prediction output of this pair.

\subsection{Target states, robot contexts, and controlled comparisons}
\label{sec:states}
The target state $q$ is a single continuous physical coordinate, such as a joint displacement, rotation angle, or object-position component. A scene $s$ is an independent evaluation unit, corresponding to one simulator initialization. Let $P$ and $K$ denote the numbers of robot contexts and target-coordinate values, with indices $p\in\{1,\ldots,P\}$ and $k\in\{1,\ldots,K\}$. They count evaluation conditions, not coordinate dimensions. We denote an observation and its measured coordinate by $o_{spk}$ and $q_{spk}$.

Robot context specifies the robot's pose and configuration at observation time; associated changes in robot appearance and object occlusion can also affect the image. Natural trajectories couple coordinate and context changes. Crossing $P$ contexts with $K$ coordinates provides two comparisons: varying $k$ at fixed $p$ evaluates responsiveness, whereas varying $p$ at fixed $k$ evaluates stability at the same state.

The fixed quantities must agree within task-specific tolerances at the final observation time. Metrics are computed on valid crossed grids specified by each analysis. A metric is not estimated if the required endpoints, coordinate separation, or context set cannot be obtained. Cells are not excluded based on prediction results.

\subsection{Accuracy, responsiveness, and context sensitivity}
\label{sec:metrics}
\paragraph{Prediction accuracy.}
We use mean absolute error (MAE). Natural MAE first averages observation errors within each rollout, then weights multiple rollouts from the same scene equally. Controlled MAE weights the crossed observations within a scene equally. Each measures accuracy under its respective observation condition.

\paragraph{Target-state responsiveness.}
For the two endpoints $k=1,K$ of the tested coordinate range, define the signed endpoint gain and its scene mean as
\begin{equation}
g_{sp}=\frac{\hat q_{spK}-\hat q_{sp1}}{q_{spK}-q_{sp1}},
\qquad G_s=\frac{1}{P}\sum_{p=1}^{P}g_{sp}.
\label{eq:gain}
\end{equation}
Both differences retain their signs. Thus, $g_{sp}=1$ indicates a prediction change with the correct direction and magnitude, zero indicates equal endpoint predictions, and a negative gain indicates the opposite direction. Gain is a dimensionless response ratio, not a distance. To detect cancellation between under- and over-response across contexts, we also measure
\begin{equation}
E_{G,s}=\frac{1}{P}\sum_{p=1}^{P}|g_{sp}-1|.
\label{eq:gainerror}
\end{equation}
An error of zero means unit endpoint gain in every tested context. Endpoint gain does not determine intermediate linearity, monotonicity, or absolute prediction error; we therefore examine MAE and coordinate-wise prediction curves alongside it.

\paragraph{Context sensitivity.}
We average the prediction range across contexts at each target coordinate:
\begin{equation}
R_s=\frac{1}{K}\sum_{k=1}^{K}\left(\max_p\hat q_{spk}-\min_p\hat q_{spk}\right).
\label{eq:range}
\end{equation}
A small $R_s$ indicates less variation across the tested contexts. A constant predictor also has $R_s=G_s=0$, so stability is interpreted jointly with responsiveness and accuracy. This diagnostic concerns the state readout; it does not require identical policy actions at different robot configurations. All metrics are computed per scene and summarized with equal scene weights. MAE and $R_s$ retain the coordinate's physical units; $g_{sp}$, $G_s$, and $E_{G,s}$ are dimensionless.

\subsection{Readout fitting and behavioral evaluation}
We freeze the feature extractor and fit readouts using training features, with normalization determined on training data. We compare full-natural training with natural and controlled training matched in sample count and actual target coordinates. Joint training specifies the contribution of each data source to the loss. Each fitted representation--readout pair remains fixed during evaluation.

Behavioral relevance is evaluated separately. Under the same observation conditions, we record readout predictions, policy actions, subsequent object states, and task success. Diagnostic outputs are not fed back to the policy. We add controlled diagnostics to failure predictors using initial state error and physical variables, then compare losses on separate evaluation initializations. A baseline using MAE from the same controlled observations distinguishes the value of additional observations from that of their diagnostic summaries.

\subsection{Controlled observation construction and validation}
Task-specific state-setting and restoration procedures establish target coordinates and robot configurations. After the specified settling or state-maintenance checks, we validate the physical state and image actually used for evaluation, including coordinate, robot configuration, object stability, velocity, and contact. Matched-training observations reproduce the measured coordinates of selected natural observations, and labels are checked after generation.

Visual checks assess target visibility, image differences induced by coordinate changes, and repeated-rendering agreement. Segmentation and physical metadata are used for validation, not as primary readout inputs. Natural observations retain their original occlusion and image distribution.

Physical validity and training support are assessed separately. Robot proximity, coordinate range and density, and joint distances over physical variables are distinct checks. Passing a limited check does not establish equality of the full state--context distribution. Support-conditioned analyses are reported as non-estimable if valid comparisons fall below their fixed minimum. Appendices~\ref{app:a}, \ref{app:e}, and~\ref{app:f} specify the procedures and thresholds.

\subsection{Evaluation protocol and statistical analysis}
Training, development, confirmation, and separate follow-up evaluations have distinct roles. Confirmation data are excluded from fitting and model or hypothesis selection; primary comparisons and statistical procedures are internally fixed before accessing these data. Additional analyses of previously examined data are identified as post hoc.

Independent initializations are the statistical units. Coordinates, contexts, frames, training seeds, and action-noise repetitions within an initialization do not increase the independent sample count. We average scene metrics across MLP seeds and preserve initialization pairing in comparisons and bootstrap resampling. Individual intervals describe evaluation-initialization variability conditional on fixed training data and fitted models.

The primary confirmation uses exact sign tests for directional consistency and paired bootstrap intervals for mean effects and differences. Test families, adjustments, tie rules, and resampling procedures are specified for each experiment. Development and follow-up results are not pooled into the primary confirmatory sample or test family.

\section{Experiments}
\label{sec:experiments}
The primary confirmation evaluates drawer displacement and faucet rotation in Meta-World~\citep{yu2020metaworld}, using ridge readouts fitted to frozen SmolVLA~\citep{shukor2025smolvla} and OpenVLA~\citep{kim2024openvla} features. Table~\ref{tab:design} distinguishes the datasets and roles of the evaluations.

\begin{table}[htbp]
\caption{Evaluation design. Behavioral experiments use a task-adapted, frozen SmolVLA policy in LIBERO~\citep{liu2023libero}, separately from the Meta-World representation analysis. Appearance interventions, stove evaluations, and development controls use the distinct sample counts reported with their results.}
\label{tab:design}
\centering\small
\begin{tabularx}{\linewidth}{@{}L{.22\linewidth}YY@{}}
\toprule Evaluation & Data and independent units & Purpose \\
\midrule
Primary confirmation & 24 train / 8 development / 12 confirmation initializations per task; 4 contexts $\times$ 6 coordinates per scene & Natural accuracy, controlled response, matched training \\
Representation/model extension & 32 new evaluation initializations per task; radial/lateral context axes & Fixed-candidate replication and comparison \\
Free object & 32 mug-position initializations; 64 support follow-up initializations & Physical scope and estimability \\
Failure prediction & 96 development / 128 evaluation initializations; 2 normal action-noise rollouts each & Additional pre-rollout outcome information \\
\bottomrule
\end{tabularx}
\end{table}

\subsection{Natural accuracy and controlled responsiveness}
\label{sec:primary}
\begin{figure}[tb]
\centering\includegraphics[width=\linewidth]{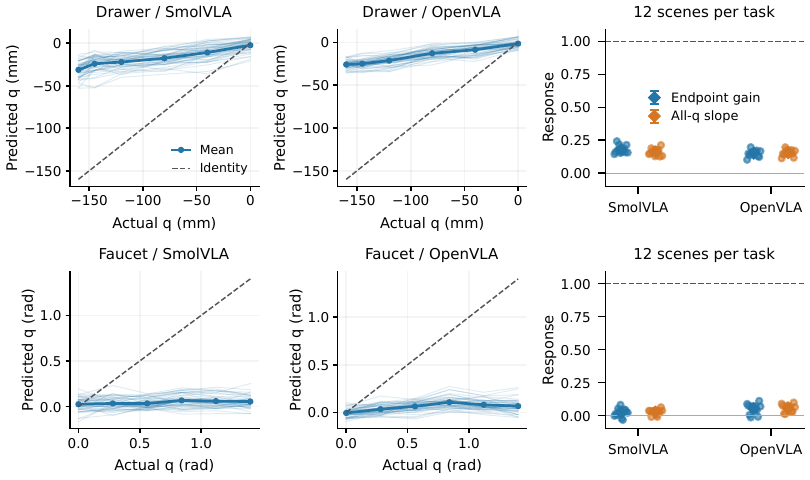}
\caption{Confirmatory predictions and responses. Left: all 12 scenes $\times$ 4 context curves per combination (faint), mean predictions (bold), and identity lines (dashed). Right: scene-level endpoint gains and post-hoc all-coordinate slopes; large markers and bars show means and individual 95\% intervals. Both summaries are below unit response. Mean gain does not characterize intermediate non-monotonicity.}
\label{fig:confirmation}
\end{figure}

We first test whether an accurate natural-trajectory readout responds faithfully to controlled physical changes. \textbf{Low natural MAE and small endpoint gain coexist in all four task--representation combinations.} Table~\ref{tab:primary} reports all three diagnostic axes, and Figure~\ref{fig:confirmation} shows every confirmatory scene's predictions and gains.

\begin{table}[htbp]
\caption{Primary confirmation with full-natural readouts. Values are equal-weight means over 12 scenes per task. MAE and $R$ use mm for Drawer and rad for Faucet; $G$ and $E_G$ are dimensionless. Brackets are individual 95\% scene-bootstrap intervals. Unit mean gain does not imply accuracy at every intermediate state. Appendix~\ref{app:a7} reports intervals for all metrics.}
\label{tab:primary}
\centering\small
\begin{tabular}{@{}llrrrrr@{}}
\toprule Task & Model & Nat. MAE & Ctrl. MAE & $G$ [95\% CI] & $E_G$ & $R$\\
\midrule
Drawer & SmolVLA & 5.554 & 74.184 & 0.1806 [0.1653, 0.1978] & 0.8194 & 16.499\\
 & OpenVLA & 3.892 & 75.884 & 0.1525 [0.1387, 0.1662] & 0.8475 & 9.797\\
Faucet & SmolVLA & 0.0763 & 0.6698 & 0.0211 [0.0035, 0.0388] & 0.9789 & 0.1285\\
 & OpenVLA & 0.0570 & 0.6475 & 0.0512 [0.0292, 0.0717] & 0.9488 & 0.1008\\
\bottomrule
\end{tabular}
\end{table}

All 12 scenes in each combination have $G_s<1$. The four responsiveness sign tests each yield Holm-adjusted $p=0.001953125$ across the eight primary hypotheses. These tests establish directional consistency; the curves and effect sizes show the response magnitude. Slopes fitted to all six coordinates are also small, showing that the low global linear response is not specific to the two-endpoint summary (Appendix~\ref{app:b7}). Natural prediction accuracy and faithful response to controlled coordinate changes are therefore distinct properties.

\subsection{Robustness to evaluation, readout, representation, and model choices}
\label{sec:robustness}
The separation persists in actual-coordinate-matched evaluations and with RBF and MLP readouts. Some nonlinear settings improve controlled error or response, but improvements in natural accuracy do not consistently improve responsiveness and stability.

We evaluate fixed representation-location, pooling, and readout candidates on new initializations, and extend the comparison to $\pi_0$ base~\citep{black2024pi0} and GR00T N1.6~\citep{nvidia2025groot}. Table~\ref{tab:models} summarizes representative gains. Frozen SigLIP~2, VC-1, R3M, and DINOv3 features exhibit the same type of separation~\citep{tschannen2025siglip2,majumdar2023vc1,nair2023r3m,simeoni2025dinov3}; the separation is therefore not specific to VLA architectures.

\begin{table}[tb]
\caption{Mean endpoint gain of fixed final-mean-ridge candidates on 32 new initializations per task. Radial and lateral axes share initializations and use local $\pm10$\,mm hand-position changes relative to the robot. Model-native input paths differ, so this is not a controlled ranking of model quality. Appendix~\ref{app:b} provides MAE and candidate details.}
\label{tab:models}
\centering\small
\begin{tabular}{@{}lrr@{}}
\toprule Model & Drawer $G$: radial / lateral & Faucet $G$: radial / lateral\\
\midrule
SmolVLA & 0.2058 / 0.2060 & 0.0097 / 0.0122\\
OpenVLA & 0.1692 / 0.1661 & 0.0429 / 0.0394\\
$\pi_0$ base & 0.1094 / 0.1086 & $-0.0106$ / $-0.0011$\\
GR00T N1.6 & 0.2413 / 0.2426 & 0.0428 / 0.0435\\
\bottomrule
\end{tabular}
\end{table}

Actual-coordinate matching within natural observations also reveals prediction variation, although it does not directly measure low endpoint responsiveness on the natural distribution. Coordinate matching and readout comparisons weaken simple coordinate-range and single-probe explanations. Section~\ref{sec:discussion} discusses the remaining joint-distribution shift.

\subsection{Scope across physical state definitions and task settings}
\label{sec:scope}
We evaluate a mug's world-$x$ position to examine the pattern beyond articulated coordinates. As Table~\ref{tab:scope} shows, the combination of low natural error and weak controlled response does not reproduce in this free-object setting. The physical structure and support of natural training coordinates affect which crossed comparisons can be constructed. Appendices~\ref{app:e} and~\ref{app:f} report book and other task-construction results.

\begin{table}[tb]
\caption{Free-object results and estimability. The first evaluation passes its physical and visual criteria. The support follow-up is non-estimable, not a zero-gain or model-response failure. The target is one position component, not full 6-DoF state.}
\label{tab:scope}
\centering\small
\begin{tabularx}{\linewidth}{@{}L{.25\linewidth}rY@{}}
\toprule Evaluation & Initializations & Result\\
\midrule
6\,cm position variation & 32 & Natural MAE 128.314\,mm; controlled MAE 39.765\,mm; $G=0.140$. Natural error is already large.\\
Support-conditioned follow-up & 64 & 43 physically valid; 5 support-valid; intersection 0. Below the minimum of 32 for estimation.\\
\bottomrule
\end{tabularx}
\end{table}

\subsection{Training interventions and diagnostic selection}
\label{sec:training}
\begin{figure}[tb]
\centering\includegraphics[width=\linewidth]{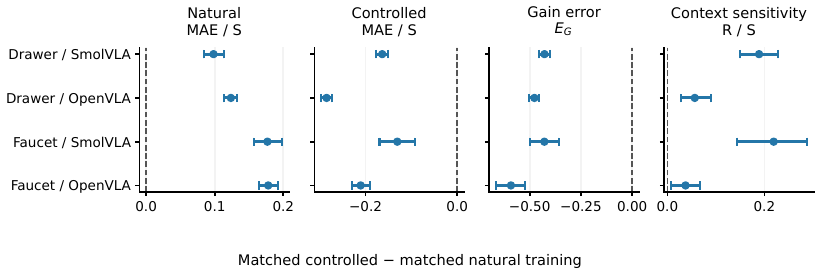}
\caption{Matched controlled minus matched natural training. Points and bars show means and individual 95\% paired scene-bootstrap intervals over 12 confirmation scenes. Negative values indicate lower error or sensitivity. For display only, MAE and $R$ are divided by $S=0.16$\,m for Drawer and $S=1.4$\,rad for Faucet; metric definitions are unchanged.}
\label{fig:training}
\end{figure}

We first replace 36 natural observations with 36 controlled observations matched in training scene, sample count, and actual coordinate. Figure~\ref{fig:training} shows lower controlled MAE and gain error but higher natural MAE and mean context sensitivity. This is the tradeoff observed in this matched replacement comparison.

In development analyses, changing coordinate--context pairing changes error and responsiveness even with fixed sample count and coordinate/context marginals. Joint natural--controlled training can reduce both MAEs without reducing context sensitivity. A 50:50 MLP mixture improves both MAEs relative to matched-natural training in all four combinations under the original evaluation, but only one combination under actual-coordinate matching. The evidence supports diagnostic changes that depend on training composition and evaluation conditions, rather than an unavoidable tradeoff.

Candidate comparisons likewise identify Pareto alternatives with advantages in gain error or context sensitivity. Candidates fixed on development data do not retain all relative advantages on new initializations. The minimum-natural-MAE candidate belongs to every development group's Pareto set, so the two criteria do not establish different unique winners. Appendix~\ref{app:c} reports the training and candidate comparisons.

\subsection{Readout diagnostics and policy behavior}
\label{sec:behavior}
Natural whole-trajectory MAE is associated with success, but intervention-condition accuracy rankings do not always preserve success rankings. At the same initial target coordinate, changing robot context reduces MAE from 38.766 to 30.684\,mm while successes decrease from 106 to 52 out of 128. This is a descriptive comparison of errors measured after rollout; context also changes geometry and the required actions.

To evaluate pre-rollout information, we fit failure predictors on 96 development initializations and evaluate two normal-noise rollouts from each of 128 separate initializations. $M_0$ uses initial state error and physical variables; the original $M_{GR}$ adds $E_G$ and $R$. Post-hoc controls add MAE from the same six controlled snapshots ($M_C$), or MAE together with $E_G,R$ ($M_{CGR}$). Existing models are not refitted; both new models are fitted only on development data.

\begin{table}[tb]
\caption{Failure-prediction losses on the same 128 evaluation initializations; lower is better. Two normal-noise rollouts are grouped within each equally weighted initialization. $M_0/M_{GR}$ form the original fixed comparison; $M_C/M_{CGR}$ are post-hoc additions. This evaluates failure prediction, not an improvement to policy success.}
\label{tab:behavior}
\centering\small
\begin{tabular}{@{}llrr@{}}
\toprule Model & Information added to $M_0$ & Brier & Log loss\\
\midrule
$M_0$ & None & 0.181515 & 0.543533\\
$M_C$ & Controlled MAE & 0.175654 & 0.528602\\
$M_{GR}$ & $E_G,R$ & 0.171381 & 0.519075\\
$M_{CGR}$ & Controlled MAE, $E_G,R$ & 0.169677 & 0.514777\\
\bottomrule
\end{tabular}
\end{table}

\begin{figure}[tb]
\centering\includegraphics[width=\linewidth]{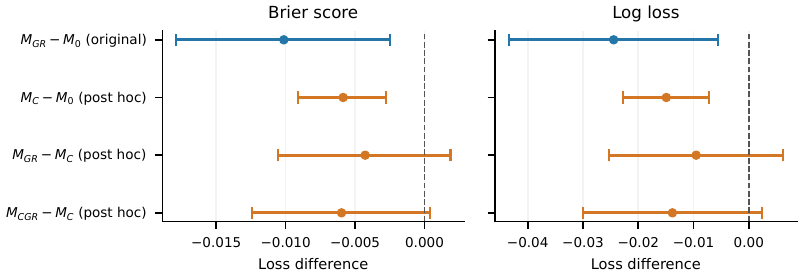}
\caption{Mean loss differences and individual 95\% intervals from 20,000 paired initialization-bootstrap draws. Negative differences favor the first model. Original and post-hoc comparisons are distinguished. Intervals describe evaluation uncertainty conditional on the development fits.}
\label{fig:behavior}
\end{figure}

\textbf{Controlled diagnostics provide additional policy-failure information beyond initial error and physical variables.} Both $M_{GR}$ and $M_C$ reduce Brier score and log loss relative to $M_0$ (Table~\ref{tab:behavior}, Figure~\ref{fig:behavior}). For comparisons between summaries of the same controlled observations, the individual 95\% intervals for $M_{GR}-M_C$ and $M_{CGR}-M_C$ include zero for both losses.

Additional evaluations distinguish readout, action, and outcome. Changing robot appearance at the same physical state changes the readout and initial action, but paired initial object-progression differences are zero and success-difference intervals include zero. On Stove, small readout responses coexist with high policy success. Understanding the behavioral meaning of a diagnostic therefore requires examining predictions, actions, physical progression, and success together. Appendix~\ref{app:f} provides the task-specific results.

\section{Discussion and Conclusion}
\label{sec:discussion}
\subsection{Discussion}
Natural state-prediction accuracy establishes that a state can be decoded under natural observations. Our results show that this evidence does not substitute for faithful response to controlled state changes or stability across contexts at the same state. The different diagnostic changes under representation, readout, and training choices indicate that selecting candidates by natural MAE alone can overlook meaningful distinctions.

Time and robot configuration alone predict target coordinates accurately on natural trajectories (Appendix~\ref{app:g4}). This establishes contextual predictability, not that the VLA uses the same pathway. Coordinate matching, visual validation, and model comparisons also leave generalization to new state--context combinations unresolved. The empirical finding is that natural accuracy and controlled responsiveness can separate in the tested fixed representation--readout pairs.

Adding controlled diagnostics improves failure prediction beyond initial error and physical variables, and controlled MAE is also informative. Controlled diagnostics, including controlled MAE, provide additional policy-failure information beyond initial state-prediction error and the specified physical variables. Responsiveness and context sensitivity additionally describe the direction and magnitude of prediction changes and their variation across contexts. Using these diagnostics to improve policy success remains a future research direction.

\subsection{Limitations and scope}
\textbf{Application to real robots requires further validation.} State--context comparisons and outcome-prediction relationships established in simulation may change under real sensor errors, contact, friction, and scene variation. How to use these diagnostics to improve actual task success remains open.

The evaluations cover single physical coordinates in a limited set of tasks and contexts. The free-object setting does not reproduce the same central pattern, and some comparisons cannot jointly satisfy physical and support criteria. Controlled observations may lie outside the natural joint distribution; the present results do not separate representation content, readout feature selection, and distribution-shift effects. Endpoint gain and $R$ also depend on the chosen coordinate range and context set.

Behavioral prediction results concern the evaluated fixed policy, task, and specified baselines. Controlled-MAE comparisons and all-coordinate slopes are post-hoc analyses, with individual intervals conditional on fixed training data and fitted models. We do not establish that the diagnostic readout is the policy's internal state estimator or a causal mediator of failure.

\subsection{Conclusion}
High state-prediction accuracy on natural images alone does not establish faithful response to controlled target-state changes or downstream task success. We demonstrate distinct accuracy, responsiveness, and context-stability properties in fixed representation--readout pairs, and show that controlled diagnostics provide additional policy-failure information beyond initial error and physical variables. VLA evaluation and design should consider these properties together and examine their relationship to policy behavior and task outcomes.

\label{main-end}
\clearpage
\section*{AI Use Statement}
We used ChatGPT and Codex to assist with drafting selected sections of the manuscript, translation and grammatical editing, preparing and editing draft figures and tables, searching for relevant literature, and executing selected experimental code and organizing its results. AI-generated outputs and interpretations were used to support the preparation and refinement of the study, including the design of follow-up analyses and experiments.

Vision-language models were also used to assist with visual inspection of selected rendered observations and figures. These assessments were used as supporting information and were reviewed by the authors.

All AI-assisted outputs, experimental results, analyses, figures, references, and manuscript text were reviewed and verified by the authors. The authors made all final scientific and editorial decisions and take full responsibility for the manuscript and its conclusions.

\section*{Reproducibility Statement}
Section~\ref{sec:method} defines the evaluation and metrics. Appendix~\ref{app:a} specifies data splits, feature extraction, normalization, fitting, validation tolerances, and statistical procedures. Appendices~\ref{app:b}--\ref{app:g} describe the additional evaluations, training interventions, visual checks, physical-setting extensions, behavioral experiments, and support analyses. Model revisions, seeds, independent evaluation units, and the distinction between confirmatory and post-hoc analyses are provided with the corresponding procedures. Appendix~\ref{app:controlled-examples} provides examples of controlled observations.

\bibliography{references}
\bibliographystyle{iclr2027_conference}
\clearpage
\appendix
\setcounter{table}{0}
\setcounter{figure}{0}
\renewcommand{\thetable}{S\arabic{table}}
\renewcommand{\thefigure}{S\arabic{figure}}
\renewcommand{\theHtable}{S\arabic{table}}
\renewcommand{\theHfigure}{S\arabic{figure}}
\section{Primary construction and statistical protocol}
\label{app:a}
This appendix specifies the primary Meta-World evaluation. Appendices~\ref{app:b}--\ref{app:g} provide robustness analyses, training comparisons, visual controls, physical scope, behavioral evaluation, and additional controls, respectively.

\subsection{Tasks, splits, and observation sampling}
We use \texttt{drawer-open-v3} and \texttt{faucet-open-v3} in Meta-World 3.1.1 with MuJoCo 3.3.0~\citep{todorov2012mujoco}, retaining the native tasks and dynamics. Each task has 24 training, eight development, and 12 confirmation initializations. Natural training contains 652 drawer and 479 faucet observations; development contains 220 and 156, respectively. Across both tasks, confirmation contains 563 natural and 576 controlled observations.

Confirmation initializations are excluded from fitting, selection, and evaluation-design development. Models, feature paths, normalization, fitting, rendering, admission criteria, hypotheses, and analyses were fixed internally before accessing the confirmation data. This was an internally fixed protocol, not external preregistration. Natural trajectories are sampled every four frames, additionally including the first successful and terminal frames without duplication. Natural occlusion is retained; the controlled-visibility filter is not applied to natural observations.

Each controlled scene crosses four robot contexts with six coordinates. Context anchors are frames 6, 12, 24, and 36 for Drawer, and 12, 16, 20, and 24 for Faucet. These indices do not imply identical poses across scenes. Drawer coordinates are $[0,-0.04,-0.08,-0.12,-0.145,-0.16]$\,m; faucet coordinates are $[0,0.28,0.56,0.84,1.12,1.40]$\,rad.

\subsection{Frozen features, normalization, and readout fitting}
Encoders remain frozen, and diagnostic predictions are not fed to a policy. The primary SmolVLA feature concatenates four spatial means of final prefix image tokens (3,840 dimensions). The primary OpenVLA feature averages 256 final-layer image hidden states (4,096 dimensions). We retain model-native preprocessing, fixed task instructions, and input paths. OpenVLA's causal image tokens cannot attend to subsequent language tokens; the models therefore do not provide identically language-conditioned features.

The diagnostic readout receives no ground-truth coordinate, numerical proprioception, scene/frame identifier, task progress, object/goal vector, or segmentation mask. Such metadata are reserved for validation and explicitly privileged controls. For full-natural training, each trajectory has equal total weight, distributed equally among its sampled observations. For features $z_i$ and weights summing to one, normalization is
\begin{equation}
\mu=\sum_iw_i z_i,\qquad a=\sqrt{\sum_iw_i\lVert z_i-\mu\rVert^2},\qquad x_i=(z_i-\mu)/a.
\end{equation}
Training conditions sharing a representation use the same full-natural normalization. Ridge regression minimizes weighted squared error plus $\lambda\lVert\beta\rVert^2$, with $\lambda=0.001$ and an unpenalized intercept.

\begin{table}[htbp]\centering\small
\caption{Primary training conditions. Implementation names are included solely to map stored artifacts to scientific conditions.}
\begin{tabularx}{\linewidth}{@{}l l Y@{}}\toprule
Condition & Artifact name & Training observations\\\midrule
Full natural & \texttt{natural\_full} & All 24 natural training trajectories\\
Matched natural & \texttt{natural36} & Six scenes, six observations each; 36 total\\
Matched controlled & \texttt{exact\_control36} & The same six scenes and 36 actual coordinates\\\bottomrule
\end{tabularx}\end{table}

Matching sample count and coordinate values does not match the joint distribution of robot configuration, contact, velocity, and appearance.

\subsection{State restoration, settling, and visual admission}
For each context, we restore the simulator state, set the native object joint coordinate and zero object velocity, and apply 60 fixed neutral actions under the original dynamics. A fixed quantity means agreement of the final measured state within tolerance, rather than agreement of nominal commands. Matched training observations reproduce the natural observation's actual coordinate; they are not relabeled grid observations.

We use \texttt{corner4} for Drawer and \texttt{corner3} for Faucet, with $256\times256$ RGB, MSAA disabled, and fixed geometry, lighting, shadows, and hidden sites. Five repeated renders are generated for each observation. Confirmation repeats are bit-identical. Target visible area must be at least 64 pixels, and endpoint centroid displacement at least four pixels. All 24 confirmation scenes and 96 scene--context groups pass without replacement. These criteria establish the intended comparison, not membership in the natural joint distribution.

\subsection{Independent units and fixed inference}
A scene is an independent simulator initialization. Frames, contexts, readouts, training seeds, and bootstrap draws do not increase the number of scenes. Scene metrics have equal weight; multiple natural rollouts are averaged equally within scene.

The eight primary hypotheses cross two tasks and two representations with (i) $1-G_s>0$ for full-natural readouts and (ii) positive controlled-MAE differences between matched-natural and matched-controlled training. One-sided exact sign tests use fixed ties and Holm correction across all eight hypotheses. They test directional consistency, not the mean effect magnitude.

We use 20,000 paired scene-bootstrap draws for individual percentile 95\% intervals of mean effects. All compared conditions retain scene pairing. Intervals condition on fixed training data and fitted pairs. Coordinate--context assignment analyses instead use 5,000 paired draws over eight development scenes; these are post-hoc, descriptive, unadjusted intervals. The two populations and inferential roles are not pooled.

\subsection{Validation and numerical reproducibility}
Physical checks cover object drift, final speed, unintended contact, repeated-state agreement, and final robot consistency across target coordinates. They do not require every state during settling to be stationary. The robot-support check requires a single natural training state to satisfy the position, arm, gripper, and orientation bounds jointly; it does not certify full joint support. Natural--controlled MAE differences consequently remain descriptive rather than estimates of a context-specific causal effect.

Independent numerical replay checks features, predictions, and metrics separately from rendering reproducibility. An early MSAA-on drawer render differed by one intensity level. That discrepancy was retained, and the final MSAA-off pipeline was fixed for training feature extraction, normalization, and evaluation before confirmation. Checkpoint revisions, initialization identifiers, and checksums specify reproducibility; internal workflow names are not scientific conditions.

\subsection{Implementation details and numerical thresholds}
\begin{table}[htbp]\centering\small
\caption{Primary simulator initialization identifiers.}
\begin{tabular}{@{}llll@{}}\toprule Task & Training & Development & Confirmation\\\midrule
Drawer & 108200--108223 & 108300--108307 & 108400--108411\\
Faucet & 110200--110223 & 110300--110307 & 110400--110411\\\bottomrule
\end{tabular}\end{table}

Coordinate matching uses one-to-one assignment minimizing squared coordinate differences, implemented with \texttt{scipy.optimize.linear\_sum\_assignment} and a fixed index order for ties. Controlled matches target the final measured natural coordinate.

Neutral actions are $[0,0,0,-1]$ for Drawer and $[0,0,0,+1]$ for Faucet, applied through the native action interface for 60 steps. Object drift and final speed are each at most $10^{-3}$ (m and m/s for Drawer; rad and rad/s for Faucet). Repeated-state component differences are at most $10^{-8}$, and final robot-state spread across target coordinates is at most $10^{-7}$. One common natural training state must match end-effector position within 0.02\,m, arm-joint RMSE within 0.15\,rad, gripper RMSE within 0.005\,m, and orientation within $10^\circ$. These thresholds were not adjusted after confirmation.

The robot consistency vector contains seven arm joints (rad), two gripper joints (m), three hand-position components (m), and four quaternion components. Drawer uses each component's maximum minus minimum across coordinate conditions; Faucet uses the maximum absolute difference from the first coordinate condition. Tolerances apply componentwise in native units, not to an aggregate physical distance mixing units.

The primary sign-test tie tolerance is $10^{-8}$. Confirmation uses 20,000 bootstrap draws and seed 113200; coordinate--context assignment uses 5,000 draws. Stored task-specific indices are shared across paired conditions. Repeated observations within a scene/root are never resampled independently.

\begin{table}[htbp]\centering\small
\caption{Frozen feature paths. Model-native token grids and preprocessing are retained.}
\begin{tabularx}{\linewidth}{@{}lYY@{}}\toprule Model & Feature & Dimension/pooling\\\midrule
SmolVLA & Final prefix image tokens & Four spatial means concatenated; 3,840 primary\\
OpenVLA & Final-layer 256 causal image states & Mean; 4,096 primary\\
$\pi_0$ base & Final native image/prefix features & Mean or adaptive $2\times2$; no numerical robot state in this feature path\\
GR00T N1.6 & Final native visual tokens & Native $9\times9$ grid; mean or adaptive $2\times2$\\
SigLIP2 & Vision pooler output & 768; revision-pinned processor\\
VC-1 & Normalized CLS & 768; official resize/crop and ImageNet normalization\\
R3M & ResNet-50 pre-FC average pool & 2,048; official preprocessing\\
DINOv3 & Post-normalization CLS & 384; official $224\times224$ preprocessing\\\bottomrule
\end{tabularx}\end{table}

The SmolVLA checkpoint SHA is \path{7cd549ac2351fb069c0ddb3c34ad2d09cfc92b56a15dccdfc2e41467aaca01eb}; OpenVLA revision is \path{47a0ec7fc4ec123775a391911046cf33cf9ed83f}. VC-1 and R3M source revisions are \path{76fe35e87b1937168f1ec4b236e863451883eaf3} and \path{b2334e726887fa0206962d7984c69c5fb09cceab}. DINOv3 uses \path{facebook/dinov3-vits16-pretrain-lvd1689m}, checkpoint revision \path{114c1379950215c8b35dfcd4e90a5c251dde0d32}, and source revision \path{6876159a11b4df116f30f667f8c9888617df0751}. Its inference is FP32 with TF32 disabled.

DINOv3 uses the common inventory of 9,955 images and two task-specific ridge heads. The accepted run produces 8,064 predictions, 384 root--condition metrics, and 16 development-root metrics; independent numerical replay has maximum absolute difference $1.221\times10^{-15}$. All four non-VLA controls use natural-only task-specific ridge heads ($\lambda=0.001$), with equal total weight per natural training root, and no encoder updates or policy rollouts.

The SmolVLA/OpenVLA extension contains 64 logical candidates and 96 fitted heads, combining intermediate/final features, mean/spatial pooling, ridge/RBF/MLP, and PCA-ridge. The $\pi_0$/GR00T extension contains 24 candidates and 40 fitted heads. GR00T's adaptive-pooling bins differ from those of SmolVLA/OpenVLA. Dimension and input-path differences preclude interpreting these comparisons as a controlled ranking of representation quality.

\subsection{Complete primary metrics and uncertainty}
\label{app:a7}
Table~\ref{tab:all-primary} provides all primary intervals using the original scene metrics and resampling indices. Recomputed predictions used for the figures agree with the stored metrics.
\begin{table}[htbp]\centering\small
\caption{Complete primary metrics: mean [individual 95\% scene-bootstrap interval], $n=12$ per task. MAE/$R$ are mm for Drawer and rad for Faucet; $G,E_G$ are dimensionless.}
\label{tab:all-primary}
\begin{tabular}{@{}llrr@{}}\toprule Task & Metric & SmolVLA & OpenVLA\\\midrule
Drawer & Natural MAE & 5.554 [4.921, 6.328] & 3.892 [3.357, 4.477]\\
 & Controlled MAE & 74.184 [72.277, 75.993] & 75.884 [74.810, 76.974]\\
 & $G$ & 0.1806 [0.1653, 0.1978] & 0.1525 [0.1387, 0.1662]\\
 & $E_G$ & 0.8194 [0.8022, 0.8347] & 0.8475 [0.8338, 0.8613]\\
 & $R$ & 16.499 [13.189, 20.415] & 9.797 [7.576, 12.035]\\\midrule
Faucet & Natural MAE & 0.0763 [0.0596, 0.0940] & 0.0570 [0.0446, 0.0707]\\
 & Controlled MAE & 0.6698 [0.6560, 0.6841] & 0.6475 [0.6326, 0.6614]\\
 & $G$ & 0.0211 [0.0035, 0.0388] & 0.0512 [0.0292, 0.0717]\\
 & $E_G$ & 0.9789 [0.9612, 0.9965] & 0.9488 [0.9283, 0.9708]\\
 & $R$ & 0.1285 [0.1115, 0.1459] & 0.1008 [0.0861, 0.1171]\\\bottomrule
\end{tabular}\end{table}

\begin{figure}[htbp]\centering
\includegraphics[width=\linewidth]{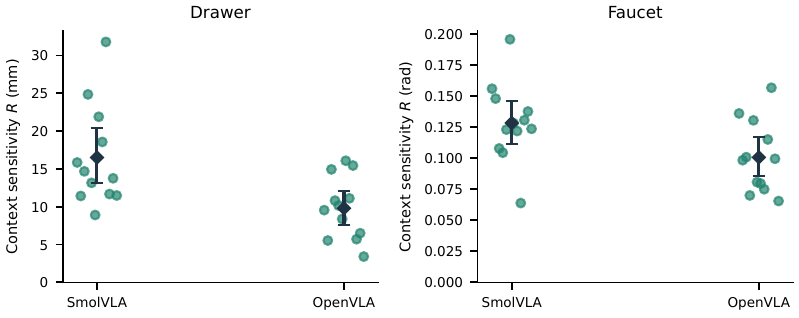}
\caption{All primary scene-level context sensitivities $R$, with means and individual 95\% intervals. Each point is one initialization; contexts and observations are not independent sample units.}
\label{fig:context}\end{figure}

\section{Robustness across evaluation, readouts, representations, and models}
\label{app:b}
\subsection{Coordinate-matched evaluation}
The primary controlled grid and natural trajectories need not share coordinate marginals. We therefore select natural development observations by measured $q$ and reconstruct the same coordinates at each controlled context. This post-hoc evaluation reuses eight development scenes per task and no confirmation scenes. Table~\ref{tab:matched-eval} uses full-natural MLPs, averaging scene metrics over three fixed training seeds rather than ensembling predictions or selecting the best seed.
\begin{table}[htbp]\centering\small
\caption{Actual-coordinate-matched development evaluation.}
\label{tab:matched-eval}
\begin{tabular}{@{}llrrr@{}}\toprule Task & Model & Matched natural MAE & Matched controlled MAE & $G$\\\midrule
Drawer & SmolVLA & 4.155\,mm & 49.369\,mm & 0.1217\\
 & OpenVLA & 3.242\,mm & 49.317\,mm & 0.1451\\
Faucet & SmolVLA & 0.1025\,rad & 0.5976\,rad & 0.0193\\
 & OpenVLA & 0.0755\,rad & 0.5873\,rad & 0.0442\\\bottomrule
\end{tabular}\end{table}

Mean errors remain separated after matching actual-coordinate marginals, with local counterexamples retained. Across the nonlinear matched analysis, controlled error is no greater than natural error in 315 of 2,304 coordinate-level rows and 1,454 of 9,216 coordinate--context rows. These repeated observations do not constitute additional independent confirmations.

\subsection{Nonlinear readouts and readout capacity}
For RBF regression, we compute the off-diagonal median $\tilde d$ of normalized training-feature distances and evaluate all nine combinations of $\sigma\in\{0.5\tilde d,\tilde d,2\tilde d\}$ and $\lambda\in\{10^{-4},10^{-3},10^{-2}\}$. The kernel is $K_{ij}=\exp(-\lVert x_i-x_j\rVert^2/(2\sigma^2))$, with an unpenalized intercept.

The MLP is \texttt{Linear(d,128)--GELU--Linear(128,1)} without dropout. Targets use training-only weighted mean/standard deviation. AdamW uses learning rate and weight decay $10^{-3}$, $\beta=(0.9,0.999)$, and $\epsilon=10^{-8}$. We run 1,000 full-batch updates without early stopping or development-based checkpoint selection, using seeds 42001, 42002, and 42003. Representative MLP values average metrics across these seeds.

\begin{table}[htbp]\centering\small
\caption{Representative full-natural development results. MAE/$R$ are mm for Drawer and rad for Faucet. RBF uses fixed $\sigma=\tilde d$, $\lambda=0.001$; MLP averages three seed-specific metrics.}
\begin{tabular}{@{}lllrrrrr@{}}\toprule Task & Model & Readout & Nat. MAE & Ctrl. MAE & $G$ & $E_G$ & $R$\\\midrule
Drawer & SmolVLA & Linear & 4.846 & 76.295 & 0.1785 & 0.8215 & 15.514\\
 & & RBF & 6.068 & 75.934 & 0.1696 & 0.8304 & 15.316\\
 & & MLP & 3.650 & 78.502 & 0.1506 & 0.8494 & 13.176\\
 & OpenVLA & Linear & 3.606 & 76.217 & 0.1487 & 0.8513 & 10.534\\
 & & RBF & 4.590 & 77.307 & 0.1262 & 0.8738 & 9.752\\
 & & MLP & 2.408 & 78.310 & 0.1337 & 0.8663 & 8.822\\\midrule
Faucet & SmolVLA & Linear & 0.0846 & 0.6625 & 0.0128 & 0.9872 & 0.1317\\
 & & RBF & 0.0892 & 0.6524 & 0.0130 & 0.9870 & 0.1207\\
 & & MLP & 0.0634 & 0.6542 & 0.0181 & 0.9819 & 0.1292\\
 & OpenVLA & Linear & 0.0623 & 0.6501 & 0.0351 & 0.9649 & 0.1159\\
 & & RBF & 0.0643 & 0.6477 & 0.0378 & 0.9622 & 0.0983\\
 & & MLP & 0.0510 & 0.6413 & 0.0405 & 0.9595 & 0.1241\\\bottomrule
\end{tabular}\end{table}

The full-natural nonlinear grid contains 48 fitted conditions: nine RBF settings and three MLP seeds for each of four task--representation combinations. Every condition has mean controlled MAE greater than natural MAE and $G<1$ over the eight development scenes. These are decoder comparisons on shared scenes, not 48 independent datasets.

RBF/MLP gain ranges are respectively 0.108--0.175/0.147--0.153 for Drawer--SmolVLA, 0.083--0.144/0.133--0.135 for Drawer--OpenVLA, 0.013--0.022/0.017--0.019 for Faucet--SmolVLA, and 0.019--0.072/0.037--0.043 for Faucet--OpenVLA. Some nonlinear readouts improve controlled error; increasing capacity does not make natural accuracy and responsiveness interchangeable. Independent prediction replay has maximum differences $8.89\times10^{-16}$ for linear, $3.27\times10^{-14}$ for RBF, and $4.12\times10^{-7}$ standardized-target units for MLP.

\subsection{Representation location, pooling, and dimensionality controls}
We cross intermediate/final features with mean/$2\times2$ spatial pooling and add 16-dimensional PCA-ridge for SmolVLA and OpenVLA. All 64 logical candidates have development mean controlled MAE greater than natural MAE and $G<1$. Pooling/readout changes can improve natural MAE while worsening $E_G$, or vice versa.

On the new evaluation bank, spatial pooling improves natural MAE in all 32 comparisons with mean pooling, and MLP improves it in all 16 comparisons with ridge. Corresponding $E_G$ changes comprise 20 decreases/44 increases for pooling and 12 decreases/20 increases for MLP. Natural MAE uses common natural observations, whereas $E_G$ is measured on two context axes, doubling the comparison count. PCA uses training-only weighted covariance and 16 principal components as a capacity control.

\subsection{Independent evaluation banks and additional model families}
We use 32 new initializations per task, with radial and lateral hand-position variations of $\pm10$\,mm about a reference configuration. Representative feature--pooling--ridge settings and statistical rules were fixed after development and technical checks, before inspecting new evaluation predictions.

During technical checks, an additional post-settling nominal joint-limit gate failed in 3,340 rows and was removed. Initial inverse-kinematics joint limits and the existing physical/visual checks were retained, as were the original failures and 230 physically invalid matched-observation rows. The modified contract was fixed before inspecting predictions. This bank should not be described as passing every original confirmation gate unchanged.

\begin{table}[htbp]\centering\small
\caption{Fixed final-mean-ridge results on 32 new initializations per task. MAEs are mm for Drawer and rad for Faucet. Paired entries denote radial/lateral axes.}
\begin{tabular}{@{}llrrr@{}}\toprule Task & Model & Natural MAE & Controlled MAE & $G$\\\midrule
Drawer & SmolVLA & 7.303 & 72.420 / 74.559 & 0.2058 / 0.2060\\
 & OpenVLA & 4.317 & 76.258 / 76.555 & 0.1692 / 0.1661\\
 & $\pi_0$ base & 6.689 & 81.547 / 82.161 & 0.1094 / 0.1086\\
 & GR00T N1.6 & 3.711 & 69.011 / 68.812 & 0.2413 / 0.2426\\\midrule
Faucet & SmolVLA & 0.0698 & 0.6867 / 0.6792 & 0.0097 / 0.0122\\
 & OpenVLA & 0.0484 & 0.6789 / 0.6778 & 0.0429 / 0.0394\\
 & $\pi_0$ base & 0.0703 & 0.6897 / 0.6881 & $-0.0106$ / $-0.0011$\\
 & GR00T N1.6 & 0.0468 & 0.6375 / 0.6392 & 0.0428 / 0.0435\\\bottomrule
\end{tabular}\end{table}

All 176 candidate--axis summaries (88 fixed candidates, two axes) have $G<1$ and controlled MAE greater than natural MAE. They share 64 roots. For the eight prespecified representative task--model comparisons, all 32/32 roots have $G<1$, with Holm-adjusted $p=1.86265\times10^{-9}$ within this separate follow-up family. These tests are not merged with the primary eight-test family.

\subsection{Local responses and context-axis effects}
Across 130,560 adjacent-coordinate rows, 38,603 responses are negative and 458 exceed unit response. Repeated rows are not independent units. Context-axis effects are not uniformly monotone and do not support a universal model ranking. Natural-coordinate matching and training-support restriction are analyzed separately: faucet results remain directionally consistent, whereas drawer support is insufficient (Appendix~\ref{app:g1}).

\subsection{Output calibration and local-response interpretation}
An additive output correction can improve MAE while leaving $G$ and $R$ unchanged because prediction differences are unchanged. An affine rescaling can move gain toward one while also changing natural MAE and $R$. These post-hoc corrections do not retrain representations or policies and cannot establish joint improvement. Local gains retain negative and greater-than-one values without clipping.

\subsection{All-coordinate response slopes}
\label{app:b7}
As a post-hoc check of the endpoint summary, we fit an intercept and slope to all six coordinates for every original confirmation scene/context:
\begin{equation}
b_{sp}=\frac{\sum_k(q_{spk}-\bar q_{sp})(\hat q_{spk}-\overline{\hat q}_{sp})}{\sum_k(q_{spk}-\bar q_{sp})^2},\qquad B_s=\frac{1}{P}\sum_p b_{sp}.
\end{equation}
Bars denote means across the six coordinate levels. We equally weight the same 12 scenes and reuse the 20,000 bootstrap draws without changing the primary hypotheses.
\begin{table}[htbp]\centering\small
\caption{Endpoint gains and all-coordinate slopes: means [individual 95\% intervals].}
\begin{tabular}{@{}llrr@{}}\toprule Task & Model & Endpoint $G$ & All-coordinate $B$\\\midrule
Drawer & SmolVLA & 0.1806 [0.1653, 0.1978] & 0.1618 [0.1477, 0.1762]\\
 & OpenVLA & 0.1525 [0.1387, 0.1662] & 0.1561 [0.1427, 0.1691]\\
Faucet & SmolVLA & 0.0211 [0.0035, 0.0388] & 0.0262 [0.0141, 0.0373]\\
 & OpenVLA & 0.0512 [0.0292, 0.0717] & 0.0550 [0.0394, 0.0702]\\\bottomrule
\end{tabular}\end{table}

Low global linear response is therefore not confined to endpoint gain. Either summary can conceal local non-monotonicity; Figure~\ref{fig:confirmation} also reports the coordinate-level curves. No additional bank or independent sample is introduced.

\section{Training interventions and candidate selection}
\label{app:c}
\subsection{Matched replacement of natural observations}
Matched training uses six scenes and 36 actual target coordinates. Replacing natural with controlled observations reduces mean controlled MAE and $E_G$, while increasing natural MAE, in all four task--representation combinations. For Drawer--SmolVLA, controlled MAE changes from 52.645 to 26.450\,mm and natural MAE from 37.376 to 53.098\,mm. Mean $R$ increases in all four combinations, with the scene-level directions shown below.
\begin{table}[htbp]\centering\small
\caption{Scene-level context-sensitivity changes under matched controlled replacement ($n=12$ per task).}
\begin{tabular}{@{}llrr@{}}\toprule Task & Model & $R$ increases & $R$ decreases\\\midrule
Drawer & SmolVLA & 12 & 0\\
 & OpenVLA & 10 & 2\\
Faucet & SmolVLA & 11 & 1\\
 & OpenVLA & 9 & 3\\\bottomrule
\end{tabular}\end{table}

Training composition changes diagnostic behavior without necessarily improving all properties together.

\subsection{Coordinate--context assignment and coupling strength}
We vary which coordinate is paired with which robot context while holding sample count and both marginals fixed. For 36 faucet samples, we compare correlated, reversed, and two balanced assignments, preserving normalization, regularization, counts, and marginal frequencies. All eight representation--baseline/balanced comparisons improve mean development natural MAE, controlled MAE, and gain error. In one SmolVLA comparison, balanced instead of correlated assignment changes natural MAE from 0.5629 to 0.4693\,rad and controlled MAE from 0.3954 to 0.2648\,rad.

Individual context-pair stability is not uniformly improved. In the same comparison, the prediction range between context anchors 16 and 20 increases from 0.2118 to 0.2242\,rad. Extending coupling to 17 assignments and applying the principle to Drawer produces non-monotone changes. Assignment effects also appear in 24 leave-one-training-scene-out configurations. No tested coupling is consistently best across all diagnostics.

\subsection{Joint natural--controlled training}
\begin{table}[htbp]\centering\small
\caption{Joint-training designs. All fit a single readout to both observation types.}
\begin{tabularx}{\linewidth}{@{}lYY@{}}\toprule Design & Composition & Purpose\\\midrule
Joint-A & 36 matched natural + 36 matched controlled = 72 & Combine observations of the same scenes/coordinates\\
Joint-B & Full natural + 36 controlled; 688 Drawer / 515 Faucet & Retain all original natural training data\\
Fixed budget & 36 total; controlled fractions $1/3,1/2,2/3$, each with two complementary assignments & Compare mixtures at fixed sample count\\\bottomrule
\end{tabularx}\end{table}

Joint-A/B use controlled loss weights 0.25, 0.50, and 0.75, with 0.50 as the central comparison. These are loss weights, not sample fractions. The Joint-B count condition weights every combined row equally, also changing the full-natural trajectory weighting. Representations, training-only normalization, and readout families remain fixed. Verification covers 184 head/conditions including baselines and fixed-budget conditions, with 28 new linear and 84 new MLP heads.

Relative to matched-natural MLP training, the 50:50 Joint-A MLP reduces both MAEs in all four combinations under the original development evaluation. For Drawer--SmolVLA, natural MAE changes from 62.048 to 55.274\,mm and controlled MAE from 46.203 to 32.520\,mm, while $R$ increases from 26.715 to 49.512\,mm. Under coordinate-matched evaluation, only Drawer--SmolVLA retains improvement in both MAEs.

Relative to full-natural training, 50:50 Joint-B generally reduces controlled MAE and $E_G$ while increasing natural MAE and $R$ across linear/MLP and both evaluations. For Drawer--SmolVLA MLP, natural MAE changes from 3.650 to 5.259\,mm, controlled MAE from 78.502 to 39.359\,mm, and $R$ from 13.176 to 39.217\,mm. Other mixture weights and uniform-row training include exceptions where both MAEs improve, such as coordinate-matched Faucet--SmolVLA. Hence the tradeoff is not inevitable. No tested A/B condition improves all four mean metrics simultaneously, within this finite mixture/readout study.

\subsection{Candidate selection by accuracy and joint diagnostics}
Selection rules are fixed on development data and evaluated on new initializations. Natural-accuracy selection retains every candidate within $10^{-8}$ of the minimum normalized natural MAE. Joint selection retains the Pareto non-dominated set for $(\mathrm{MAE}_{\mathrm{nat}}/S,E_G,R/S)$, where $S=0.16$\,m for Drawer and $1.4$\,rad for Faucet.

Development Pareto-set sizes are 9 for each Drawer--SmolVLA/OpenVLA group, 7 for Faucet--OpenVLA, and 4 for Faucet--SmolVLA. They are 3 and 4 for Drawer--GR00T/$\pi_0$, and 3 for each Faucet--GR00T/$\pi_0$ group. Many candidates remain non-dominated on new initializations, but not every relative advantage persists.

All eight development groups include a minimum-natural-MAE candidate in their Pareto set. Different sets therefore do not establish different unique winners. We examine alternative candidates' $E_G,R$ advantages and their persistence. Evaluation Pareto sets are descriptive, not used to reselect a winner. Policy success improvement is not evaluated here.

\section{Visual discriminability and supplementary visual controls}
\label{app:d}
\subsection{Differences after image preprocessing}
We separately check whether target-state differences survive each model's actual image preprocessing. Endpoint pairs admitted in raw RGB are processed through the native resizing, cropping, and normalization path, after which target-region spatial differences and endpoint ordering are inspected. Features or predictions are not used to retain observations with a desired response. This audit is distinct from physical validity and readout responsiveness: visual distinguishability does not establish how a representation responds.

\subsection{Controlled observation examples}
Figure S2 in Appendix H shows six target-coordinate settings at one fixed robot context for each task. The observations are ordered by their measured simulator coordinate, with coordinate values and units displayed. A fixed 180-degree orientation correction presents the rendered images upright. These examples illustrate the controlled comparison; quantitative physical and visual checks remain those specified in Appendix A and Sections D.1 and D.3.

\subsection{Target-pixel preservation and rendering controls}
Primary visual admission requires at least 64 visible target pixels and at least four pixels of endpoint centroid displacement. Five renders of each state test reproducibility; all 5,695 repeated RGB comparisons in confirmation are bit-identical. An early MSAA-on one-intensity-level mismatch is retained separately. The final primary pipeline uses MSAA-off consistently for training features and evaluation.

Appearance interventions preserve the target label and physical state while changing robot or background pixels inside specified masks. Pixel-level audits check target preservation and record unintended changes outside masks. These interventions isolate specified appearance factors; they do not hold every aspect of visual context fixed.

\subsection{Privileged region pooling and visual-feature controls}
Segmentation-based object-region pooling is a privileged control, not the primary method. We compare object ROIs, equal-area control ROIs, and training-fixed ROIs on the same frozen token grid.
\begin{table}[htbp]\centering\small
\caption{Representative natural-full region-pooling controls. MAE units are mm for Drawer and rad for Faucet. Blank task/model cells inherit the preceding entry.}
\begin{tabular}{@{}lllrrr@{}}\toprule Task & Model & Pooling & Natural MAE & Controlled MAE & $G$\\\midrule
Drawer & SmolVLA & Full mean & 7.501 & 72.656 & 0.200\\
 & & Object ROI & 10.038 & 42.553 & 0.521\\
 & & Equal-area control & 6.736 & 85.426 & 0.108\\
 & OpenVLA & Full mean & 3.606 & 76.217 & 0.149\\
 & & Object ROI & 5.708 & 41.143 & 0.514\\
 & & Equal-area control & 3.586 & 70.235 & 0.243\\\midrule
Faucet & SmolVLA & Full mean & 0.0964 & 0.6976 & $-0.001$\\
 & & Object ROI & 0.1150 & 0.5291 & 0.125\\
 & & Equal-area control & 0.0965 & 0.6829 & $-0.005$\\
 & OpenVLA & Full mean & 0.0623 & 0.6501 & 0.035\\
 & & Object ROI & 0.0568 & 0.5957 & 0.057\\
 & & Equal-area control & 0.0814 & 0.4900 & 0.249\\\bottomrule
\end{tabular}\end{table}

Object ROIs substantially improve controlled error and response in some settings, but are not uniformly best across metrics. A separate bounding-box-only privileged coordinate diagnostic has natural-full gain approximately 1.071 for Drawer and 0.921 for Faucet. Explicit geometric location can therefore predict $q$; this does not imply that VLA tokens use that information in the same way.

\section{Physical-setting scope and free-object analyses}
\label{app:e}
\subsection{Free-object construction and validation}
We use a mug's world-$x$ position as the scalar target beyond articulated joints. This is a diagnostic of one position component, not the full six-degree-of-freedom pose. An initial 2\,cm translation setting supports behavioral rollouts but yields too few roots passing the fixed visual gate. We separately construct a 6\,cm setting to obtain measurable visual state changes, retaining the earlier failures. Final world-$x$, post-translation stability, unintended contact, visibility, and repeated rendering are checked. Joint-specific articulated settling rules are not transferred unchanged.

\subsection{Mug state-readout evaluation}
\begin{table}[htbp]\centering\small
\caption{Free-object results with 6\,cm state variation. Both evaluations pass their physical and visual criteria. Natural error is already large.}
\begin{tabular}{@{}lrrrr@{}}\toprule Evaluation & Roots & Natural MAE (mm) & Controlled MAE (mm) & $G$\\\midrule
Development & 16 & 121.767 & 42.293 & 0.167\\
New roots & 32 & 128.314 & 39.765 & 0.140\\\bottomrule
\end{tabular}\end{table}

Initial natural error in the 16-root evaluation is 32.826\,mm. Of 32 new-root natural rollouts, six achieve native success and 26 time out. Natural error exceeds controlled error in 27 roots; five show the reverse. Although endpoint response is small, the central low-natural-error/weak-response pattern does not reproduce because natural error is large. We therefore do not count these results as additional articulated-pattern replications.

\subsection{Support-aware free-object and additional-task analyses}
Before inspecting predictions, a free-object follow-up defines eligibility by both natural-training support and physical validity. Of 64 new candidate roots, support intervals can be calculated for 53; 159 physical state levels are checked.
\begin{table}[htbp]\centering\small
\caption{Free-object follow-up eligibility. No root satisfies both required gates.}
\begin{tabular}{@{}lr@{}}\toprule Gate & Roots\\\midrule
Physically valid & 43\\
Prespecified support-valid & 5\\
Physical and support-valid & 0\\
Required minimum for response estimation & 32\\\bottomrule
\end{tabular}\end{table}

With an empty intersection, no readout forward pass, $G$, $E_G$, or bootstrap interval is computed. Support thresholds, translation span, and target coordinate are not relaxed to acquire a sample.

Separate storage/shelf tasks evaluate book orientation and the right book's relative-$y$ coordinate. Each bank contains 32 initializations with $3\times3$ crossed observations; all 288 controlled observations lie outside that bank's prespecified support. Cached predictions permit some descriptive checks, but not a common comparison across every model, so they are not main quantitative results.

A separate book world-$x$ preparation check uses 16 initializations and ten conditions. In one initialization, every condition drifts approximately 1.20--2.15\,m after 20 zero-control steps. Passing static construction checks therefore does not guarantee state maintenance. These ten conditions count as one initialization failure. The planned 320 book policy rollouts from this construction were not run. Appendix~\ref{app:f8} reports other natural-rollout explorations.

\subsection{Boundary cases and non-estimable support}
A separate 64-root support-screened OOD bank contains 192 groups and 576 observations, all outside the q99 support bound. No root is eligible for primary estimation; no new model forward pass, readout fit, prediction, or gain is produced. The support vector comprises target $q$, nine robot joints, hand position, and quaternion. We use nearest-neighbor RMS distance after natural-training standardization, with leave-one-training-initialization-out q95 primary and q99 secondary thresholds. Coordinates are the original $q$ and $\pm0.05$ training IQR, with minimum spacing 0.001. Being outside even q99 also makes the q95 analysis non-estimable.

We distinguish physically invalid observations, physically valid observations outside the prespecified empirical support, and estimable observations satisfying physical, support, and minimum-sample criteria. Non-estimability is not encoded as $G=0$, a policy failure, or a measured negative response. Appendix~\ref{app:g1} applies the same distinction to articulated tasks.

\section{Policy evaluation and supplementary behavioral results}
\label{app:f}
\subsection{Policy, inputs, and paired execution}
Behavioral experiments use a task-adapted, frozen SmolVLA policy in LIBERO, separately from the Meta-World representation study. The policy receives RGB, robot state, and a language instruction. A diagnostic readout predicts the target coordinate from saved policy representations; its output is not fed back to the policy. We distinguish the diagnostic prediction $\hat q$, the policy action command, subsequent physical coordinate $q$, and native outcome. Simultaneous prediction/action changes do not establish that the policy uses the diagnostic readout internally.

Appearance comparisons share initial physical state and action noise. Physical-context interventions change specified robot joints and are a different design. Comparisons retain root/noise pairing; conditions and action steps are not independent samples.

The drawer task is LIBERO-90's \path{KITCHEN_SCENE2_open_the_top_drawer_of_the_cabinet_demo}, with instruction ``open the top drawer of the cabinet.'' We freeze a 40,000-step task-adapted SmolVLA checkpoint, SHA256 \path{e67afd358fb364d8fe9255e11beb935531c07f1cd81f175ff09c5c26f6e206ed}. It uses $128\times128$ agentview/wrist RGB, robot state, and language, at 20\,Hz with replanning every four actions. The horizon is 300 actions and chunk length 16. Native success requires drawer $q<-0.14$\,m. The natural-prefix ridge diagnostic aggregates $32\times960$ features into 7,680 dimensions.

Controlled snapshots cross $q_0=0$, $q_1=-0.06$\,m with first-arm-joint offsets $c_0=-0.12$, $c_1=0$, $c_2=+0.12$\,rad. Velocities are zeroed and controller targets reset. A four-zero-action check requires drawer drift at most 1\,mm, after which the intervention state is restored before snapshot acquisition and policy execution. Coordinate error, matched robot state, and restoration error use tolerance $10^{-8}$; other qpos components use $10^{-10}$. Initial negative-distance robot--target contacts are absent, and repeated RGB agreement is checked.

\subsection{Readout--action correspondence and physical progression}
The drawer appearance follow-up has 32 roots, two action-noise realizations, and six conditions (384 branches; 16,090 decisions).
\begin{table}[htbp]\centering\small
\caption{Initial changes relative to sham under identical physical state. Translation shifts are normalized command $L_2$ differences, not distances in physical coordinate units.}
\begin{tabular}{@{}lrr@{}}\toprule Condition & Mean $|\Delta\hat q|$ (mm) & First translation-command shift\\\midrule
Robot-low & 4.482 & 0.03613\\
Robot-high & 15.506 & 0.04550\\\bottomrule
\end{tabular}\end{table}

Sham exactly matches original initial predictions/actions and paired successes. Both background conditions have zero initial prediction/action shift; this is not a claim of trajectory-wide invariance. Paired coordinate-progression differences after four and 16 actions are zero for every root. Initial command changes therefore do not establish immediate object-state changes. Commands and physical coordinates have different units and are not directly subtracted.

\subsection{Additional predictive value for task outcomes}
\label{app:f3}
We fit pre-rollout failure predictors on 96 development initializations, freeze coefficients, and evaluate 128 separate initializations. Two normal-noise rollouts per evaluation root yield 175 successes and 81 failures among 256 rollouts. Development has 58 failures among 192 rollouts. The target is failure $=1$.
\begin{table}[htbp]\centering\small
\caption{Failure-predictor inputs, all measured before rollout.}
\begin{tabularx}{\linewidth}{@{}lYY@{}}\toprule Input & Definition & Scaling\\\midrule
$E_0$ & Absolute prediction error at the normal initial observation & Divide by 0.14\,m\\
$D_q$ & Initial coordinate distance to success boundary $-0.14$\,m & Divide by 0.14\,m\\
$D_{\mathrm{eef}}$ & Euclidean distance from end effector to cabinet origin & m\\
$D_{\mathrm{robot}}$ & Standardized RMS difference of seven arm joints from policy-training state mean & Policy-training standard deviation; dimensionless\\
$E_G$ & Mean context-wise endpoint gain error from six controlled snapshots & Dimensionless\\
$R$ & Prediction range over three contexts, averaged over two coordinates & Divide by 0.14\,m\\\bottomrule
\end{tabularx}\end{table}

$M_0$ uses $E_0$ and the three physical variables; the original augmented model, denoted $M_{GR}$, adds $E_G,R$. Columns use population mean/standard deviation over 96 development roots; columns with standard deviation at most $10^{-12}$ are removed. In practice $D_q=1$ throughout development and is dropped. Whole-trajectory MAE, contact, termination, and outcome information are not predictor inputs.

We average the two noise losses within root, equally weight roots, and add $0.1/2$ times the squared slope norm to logistic loss, leaving the intercept unpenalized. Fitting uses float64, zero initialization, and L-BFGS-B with maxiter 10000, gtol $10^{-10}$, ftol 0, and maxls 50. Final gradient infinity norm is at most $10^{-8}$.
\begin{table}[htbp]\centering\small
\caption{Original fixed failure-prediction comparison. Intervals condition on development fitting.}
\begin{tabular}{@{}lrrrr@{}}\toprule Loss & $M_0$ & $M_{GR}$ & Difference & Individual 95\% interval\\\midrule
Brier & 0.181515 & 0.171381 & $-0.010134$ & [$-0.017898$,$-0.002478$]\\
Log loss & 0.543533 & 0.519075 & $-0.024458$ & [$-0.043453$,$-0.005586$]\\\bottomrule
\end{tabular}\end{table}

Evaluation likewise groups both noises within root. We use 20,000 paired bootstrap draws, PCG64 seed 943122, and linear-percentile individual 95\% intervals. Brier uses raw probabilities; log loss clips at $10^{-12}$ and $1-10^{-12}$. Intervals reflect new-root sampling variation conditional on the development fits, not separate causal contributions of each diagnostic. Artifact variables named G and C correspond to $E_G$ and $R/(0.14\,\mathrm{m})$, respectively, not signed gain. Appendix~\ref{app:f7} compares controlled MAE from the same snapshots.

\subsection{Behavioral evaluation across manipulation tasks}
\begin{table}[htbp]\centering\small
\caption{Post-hoc relationship between whole-trajectory error and success, grouping two normal rollouts per evaluation root.}
\begin{tabular}{@{}lrr@{}}\toprule Successes among two rollouts & Roots & Mean whole-trajectory MAE (mm)\\\midrule
0/2 & 28 & 55.075\\
1/2 & 25 & 35.295\\
2/2 & 75 & 6.701\\\bottomrule
\end{tabular}\end{table}

Natural accuracy is associated with success. The post-hoc Pearson correlation between root-level MAE and success fraction is $r=-0.8075$; it is descriptive, not confirmatory. Rankings need not agree: at fixed $q_1$, changing $c_1$ to $c_2$ reduces whole-trajectory MAE from 38.766 to 30.684\,mm but changes success from 106/128 to 52/128. Context also changes interaction geometry and required actions, preventing a causal interpretation in terms of prediction error.
\begin{table}[htbp]\centering\small
\caption{Appearance-intervention successes out of 64 rollouts per task/condition (32 roots, two noises).}
\begin{tabular}{@{}lrrrrrr@{}}\toprule Task & Original & Sham & Robot-low & Robot-high & BG-low & BG-high\\\midrule
Drawer & 41 & 41 & 41 & 36 & 42 & 41\\
Stove & 60 & 60 & 61 & 60 & 60 & 60\\\bottomrule
\end{tabular}\end{table}

Robot-high minus sham success is $-7.8125$ percentage points for Drawer, with 95\% root-bootstrap interval [$-20.3125$,4.6875], and zero for Stove, with interval [$-4.6875$,4.6875]. Individual 98.75\% intervals for the prespecified four-comparison family are [$-23.4375$,7.8125] and [$-6.25$,6.25], respectively. Intervals including zero do not establish equivalence or absence of an effect. Stove background interventions are rendered-pixel no-ops and provide no evidence of background robustness.

\subsection{Stove development and sensitivity analyses}
\begin{table}[htbp]\centering\small
\caption{Initial development-pilot successes out of eight under different robot contexts.}
\begin{tabular}{@{}lrrr@{}}\toprule Task & Central & $-0.12$\,rad & $+0.12$\,rad\\\midrule
Drawer & 8 & 3 & 1\\
Stove & 8 & 8 & 8\\\bottomrule
\end{tabular}\end{table}

Matched-natural readouts have mean gains 0.1221 for Drawer and 0.0110 for Stove. Small stove diagnostic response coexists with high policy success. A separate stove extension has 1,216 branches from 16 development roots, with 1,128 successes and 88 timeouts. The same fixed readout has gains $-0.012517$ over the original 0--0.25\,rad interval and $-0.002556$ over 0.25--0.45\,rad. These settings share 16 roots and are not different readouts or 1,216 independent initializations.

\subsection{Intervention follow-up and path separation}
An early behavioral follow-up uses ten roots and two noises (20 unique normal trials), recording both normal and planned release-intervention branches. Forty branches are produced, but neither release triggers nor actual overrides occur. No paired treatment effect can be estimated; identical normal/release branches do not establish a zero intervention effect.

Normal trials yield 13 successes and seven failures. At the final prediction (step 296), all seven failures have actual opening 0\,mm but predicted opening 96.541--125.320\,mm. Mean whole-trajectory MAE is 71.968\,mm for failures and 6.391\,mm for successes. A single pre-action observation at $t=40$ reverses this ordering: errors are 2.996 and 5.924\,mm. The temporal summary therefore matters. Canonical finger contact occurs in 13/20 trials, all successful; restricting analysis to contact-reaching trials structurally excludes all failures.

Across these experiments, readouts diagnose observations and do not modify policy actions. The study evaluates relations among readouts, actions, physical states, and outcomes, rather than the performance of a diagnostic-guided controller.

\subsection{Same-observation controlled-MAE baseline}
\label{app:f7}
This additional analysis was specified after observing the original evaluation results and does not inherit the original $M_0$--$M_{GR}$ comparison's fixed-protocol status. Controlled MAE averages absolute errors equally over the same six pre-rollout snapshots used for $E_G,R$, then divides by 0.14\,m. $M_C$ adds this value to $M_0$; $M_{CGR}$ additionally includes $E_G,R$.

Existing $M_0/M_{GR}$ coefficients are retained. Only $M_C/M_{CGR}$ are fitted on the original 96 development roots, using the same procedure. New coefficients, normalization, code, and development-input hashes are saved before reading the evaluation inputs, without evaluation-based tuning. The overall comparison nevertheless remains post-hoc.
\begin{table}[htbp]\centering\small
\caption{All failure-prediction models on 128 evaluation roots.}
\begin{tabular}{@{}llrr@{}}\toprule Model & Added to $M_0$ & Brier & Log loss\\\midrule
$M_0$ & None & 0.181515 & 0.543533\\
$M_C$ & Controlled MAE & 0.175654 & 0.528602\\
$M_{GR}$ & $E_G,R$ & 0.171381 & 0.519075\\
$M_{CGR}$ & Controlled MAE, $E_G,R$ & 0.169677 & 0.514777\\\bottomrule
\end{tabular}\end{table}

\begin{table}[htbp]\centering\small
\caption{Paired loss differences [individual 95\% intervals], sharing 20,000 root-resampling indices.}
\begin{tabular}{@{}lrr@{}}\toprule Comparison & Brier difference & Log-loss difference\\\midrule
$M_{CGR}-M_C$ & $-0.005977$ [$-0.012421$,0.000381] & $-0.013824$ [$-0.030022$,0.002338]\\
$M_{GR}-M_C$ & $-0.004273$ [$-0.010564$,0.001869] & $-0.009527$ [$-0.025286$,0.006211]\\
$M_C-M_0$ & $-0.005861$ [$-0.009132$,$-0.002744$] & $-0.014932$ [$-0.022798$,$-0.007270$]\\
$M_{GR}-M_0$ & $-0.010134$ [$-0.017898$,$-0.002478$] & $-0.024458$ [$-0.043453$,$-0.005586$]\\\bottomrule
\end{tabular}\end{table}

Original scores and intervals are reproduced. We reconstruct $E_G,R$ and MAE from the six snapshots, check new-fit scalar gradients and probabilities, and verify intervals using a resampling-frequency matrix. No simulator rollouts or VLA fitting/inference are added.

Both loss-difference intervals for $M_C-M_0$ lie below zero. The $M_{CGR}-M_C$ and $M_{GR}-M_C$ point estimates are negative, but their intervals include zero for both losses. Controlled summaries thus add information beyond the specified initial baseline; superiority of $E_G,R$ over controlled MAE is not established. Uncertainty about differences is not equivalence. No new $p$-values or evaluation-based selection are used.

\subsection{Other task and construction outcomes}
\label{app:f8}
\begin{table}[htbp]\centering\small
\caption{Exploratory natural rollouts of fixed pick-and-place policies. Numerical errors with unknown outcomes are not imputed as failures.}
\begin{tabularx}{\linewidth}{@{}Yrrrrr@{}}\toprule Task & Roots & Planned & Success & Timeout & Unknown\\\midrule
Mug to plate & 32 & 64 & 13 & 51 & 0\\
Book to compartment & 32 & 64 & 2 & 57 & 5\\
Middle book to shelf & 32 & 64 & 1 & 48 & 15\\
Right book to shelf & 32 & 64 & 11 & 44 & 9\\\midrule
Descriptive total & 128 & 256 & 27 & 200 & 29\\\bottomrule
\end{tabularx}\end{table}

Separate mug behavioral data contain 320 rollouts from 16 roots: 30 successes, 290 timeouts, and no numerical failures. They differ from the 32 new-root state-readout evaluation in Appendix~\ref{app:e}. Book state-maintenance failures are reported there. These exploratory/construction samples are not pooled with the 128-root failure evaluation or primary confirmation.

\section{Additional support, intervention, and baseline analyses}
\label{app:g}
\subsection{Coordinate support and estimability}
\label{app:g1}
Physical validity and empirical support are recorded separately on the new initialization bank. The main six-coordinate/context grid is constructible for all 32 roots per task; matched-coordinate and support-restricted subsets differ.
\begin{table}[htbp]\centering\small
\caption{Eligible roots for each analysis, per context axis.}
\begin{tabularx}{\linewidth}{@{}Yrr@{}}\toprule Subset & Drawer & Faucet\\\midrule
Main controlled grid & 32/32 & 32/32\\
Natural-$q$-matched grid & 0/32 & 18/32\\
Target-$q$-support-restricted main grid & 0/32 & 32/32\\
Support-restricted natural-matched grid & 0/32 & 0/32\\\bottomrule
\end{tabularx}\end{table}

Mean $G<1$ and positive controlled-minus-natural MAE persist across fixed candidates in the separate 18-root matched and 32-root support-restricted faucet analyses. Support restriction leaves no eligible natural-matched grids.

This support check concerns the training range/density of target $q$, not the joint image--robot distribution. Training coordinates are divided into 20 bins, requiring at least eight observations from three training initializations. Drawer has only two supported grid coordinates, below the prespecified minimum of three distinct values. Non-estimable gains are not recorded as zero.

\subsection{State--context appearance interventions}
Robot/background pixels are changed inside fixed masks while target pixels are preserved. The resulting endpoint gains are shown below.
\begin{table}[htbp]\centering\small
\caption{Appearance effects on endpoint gain under a common mask.}
\begin{tabular}{@{}llrr@{}}\toprule Task & Model & Sham $G$ & Robot-high $G$\\\midrule
Drawer & SmolVLA & 0.2124 & 0.1730\\
 & OpenVLA & 0.1608 & 0.1126\\
 & $\pi_0$ & 0.0992 & $-0.1009$\\
 & GR00T & 0.2388 & 0.2326\\
Faucet & SmolVLA & 0.0135 & 0.0042\\
 & OpenVLA & 0.0345 & $-0.0084$\\
 & $\pi_0$ & $-0.0010$ & $-0.0837$\\
 & GR00T & 0.0474 & 0.0612\\\bottomrule
\end{tabular}\end{table}

Robot appearance often reduces response, with exceptions such as GR00T--Faucet. For Drawer robot-high, mean absolute natural-prediction shifts are 11.257\,mm for SmolVLA, 14.694\,mm for OpenVLA, 13.878\,mm for $\pi_0$, and 16.555\,mm for GR00T. These quantify sensitivity to specified appearance interventions, not a causal decomposition of every visual-context effect.

\subsection{Natural-only matched-coordinate analysis}
We post-hoc match frames within natural trajectories at nearly equal actual coordinates and measure prediction differences. No new rendering intervention is introduced. Pairs share eight development roots per task and do not increase independent sample size.

Each non-VLA model uses 9,649 drawer pairs and 2,886 faucet pairs. For DINOv3, pair-weighted mean actual-coordinate differences are 0.105\,mm and $2.91\times10^{-5}$\,rad, whereas equally root-weighted mean prediction differences are 6.390\,mm and 0.0757\,rad. Their aggregation units differ; these means are not paired effect estimates. In the VLA analysis, SmolVLA scene-mean absolute prediction differences are approximately 5.05\,mm and 0.0769\,rad. Prediction variation is therefore not confined to artificially constructed images. These observational matches do not jointly hold contact, occlusion, progress, and robot configuration fixed, and do not estimate a specific context factor's causal effect.

\subsection{Non-target contextual baselines}
\label{app:g4}
Simple time/robot-state baselines test the predictive structure of non-target variables in natural trajectories. Each task uses 24 training, eight development, and 32 evaluation roots; 12 heads produce 10,563 predictions. Combined time/configuration baselines obtain whole-trajectory MAE 1.494\,mm for Drawer and 0.0488\,rad for Faucet. Thus, progress and robot variables can strongly predict the coordinate in natural data. This does not establish that VLA features use those exact scalar variables.

\subsection{Non-VLA frozen-vision controls}
SigLIP2, VC-1, R3M, and DINOv3 use frozen visual encoders with natural-only task-specific ridge readouts. No encoder update or policy rollout is performed. The common bank has 32 roots per task. Here natural MAE uses three selected snapshots per root, rather than whole trajectories.
\begin{table}[htbp]\centering\small
\caption{Native-context gains for non-VLA controls on the common evaluation bank.}
\begin{tabular}{@{}lrr@{}}\toprule Encoder & Drawer $G$ & Faucet $G$\\\midrule
SigLIP2 & 0.3357 & 0.0794\\
VC-1 & 0.1979 & 0.0503\\
R3M & 0.3445 & 0.1391\\
DINOv3 & 0.1837 & 0.1611\\\bottomrule
\end{tabular}\end{table}

\begin{table}[htbp]\centering\small
\caption{DINOv3 native-context metrics. MAE/$R$ are mm for Drawer and rad for Faucet.}
\begin{tabular}{@{}lrrrrr@{}}\toprule Task & Natural MAE & Controlled MAE & $G$ & $E_G$ & $R$\\\midrule
Drawer & 3.947 & 77.460 & 0.1837 & 0.8163 & 7.158\\
Faucet & 0.0414 & 0.5532 & 0.1611 & 0.8389 & 0.0937\\\bottomrule
\end{tabular}\end{table}

DINOv3 uses the same image inventory and training-only protocol as the other encoders. Native and appearance predictions from the accepted run are stored and independently replayed.
\begin{table}[htbp]\centering\small
\caption{DINOv3 appearance-condition metrics, with mm for Drawer MAE/$R$ and rad for Faucet MAE/$R$.}
\begin{tabular}{@{}llrrrr@{}}\toprule Task & Appearance & Natural MAE & Controlled MAE & $G$ & $R$\\\midrule
Drawer & Original/sham & 3.947 & 77.460 & 0.1837 & 7.158\\
 & Robot-low & 12.127 & 69.297 & 0.2036 & 10.336\\
 & Robot-high & 17.906 & 55.802 & 0.2587 & 14.443\\
 & Background-low & 6.396 & 79.937 & 0.1957 & 8.143\\
 & Background-high & 8.431 & 80.598 & 0.2106 & 6.352\\\midrule
Faucet & Original/sham & 0.0414 & 0.5532 & 0.1611 & 0.0937\\
 & Robot-low & 0.0720 & 0.4730 & 0.1595 & 0.0909\\
 & Robot-high & 0.0868 & 0.5203 & 0.1764 & 0.0994\\
 & Background-low & 0.0582 & 0.5747 & 0.1393 & 0.0831\\
 & Background-high & 0.1352 & 0.6319 & 0.1325 & 0.0756\\\bottomrule
\end{tabular}\end{table}

Appearance changes can improve controlled MAE or gain while worsening natural MAE or $R$. The evidence does not identify a common causal mechanism across all models. The natural-accuracy/controlled-response separation is not uniquely specific to VLA policy architectures. These controls compare diagnostic behavior, not policy performance or a model-quality ranking.

\subsection{Cross-model and cross-task coverage}
Representation/readout factorials, additional VLAs, non-VLA encoders, and appearance interventions share or partially overlap roots and image banks. Candidate, fitted-head, and prediction counts are not counts of independent replications. The 64-root VLA bank contains 88 fixed candidates, 136 fitted heads, and 250,172 checked predictions. The first three non-VLA families each use 9,955 image-feature rows and two natural-only task heads; DINOv3 uses the same inventory and split. These counts indicate numerical coverage without increasing independent roots.

Tokenization, visual grids, language access, feature dimension, and preprocessing differ across models. Their metric differences cannot isolate a single architectural component. Model extensions test dependence on one model/path rather than rank architectures.

\subsection{Readout--policy path separation}
In Figure~\ref{fig:framework} and Appendix~\ref{app:f}, diagnostic predictions do not return to the policy. Readout/action changes and diagnosis/success associations are analyzed together without assuming the policy internally uses the fitted readout. The additional outcome comparison retains this separation.

\subsection{Alternative explanations and OOD scope}
\begin{table}[htbp]\centering\small
\caption{Alternative explanations and the scope of the available controls.}
\begin{tabularx}{\linewidth}{@{}YYY@{}}\toprule Explanation & Available evidence & Remaining question\\\midrule
Physical/rendering errors & Actual coordinate, robot, visibility, repeated-render and target-pixel checks & Does not establish natural joint-distribution membership\\
Different target-coordinate ranges & Actual-coordinate matching and separate faucet state-value matching & Does not match the entire context distribution\\
Variation confined to artificial images & Prediction differences in natural-only near-equal-$q$ pairs & Observational matching does not isolate a causal context factor\\
One VLA or readout fails & Multiple VLAs, feature paths, pooling, nonlinear readouts, non-VLA controls & Models may share sensitivity to distribution shift\\
Novel state--context combinations are jointly OOD & Limited support-aware follow-ups & Joint-distribution OOD remains possible\\\bottomrule
\end{tabularx}\end{table}

A separate faucet state-value-matched follow-up uses a post-hoc restricted set of four initializations. After matching actual $q$, four VLA models have mean gains approximately 0.016--0.070 and controlled-minus-matched-natural MAE approximately 0.470--0.514\,rad. This weakens a coordinate-mismatch-only explanation but is not in-distribution confirmation of the full joint state--context distribution.

The controls make target-coordinate range, simple image-construction errors, a single readout, or a single VLA insufficient as complete explanations. Generalization failure on state--context combinations outside the natural joint distribution remains possible. We therefore interpret the findings as a diagnostic separation between natural accuracy and responsiveness under validated interventions, rather than as identification of an OOD-independent mechanism.

\clearpage
\section{Controlled observation examples}
\label{app:controlled-examples}
\begin{figure}[h]\centering
\setlength{\tabcolsep}{2pt}
\textbf{Drawer}: native joint coordinate (mm)\par\smallskip
\begin{tabular}{@{}cccccc@{}}
$-160$ & $-145$ & $-120$ & $-80$ & $-40$ & $0$\\
\includegraphics[angle=180,width=0.158\linewidth]{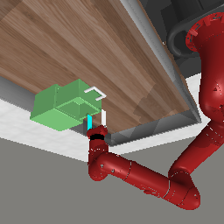} & \includegraphics[angle=180,width=0.158\linewidth]{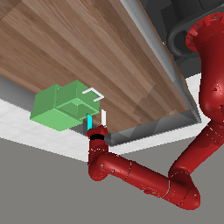} & \includegraphics[angle=180,width=0.158\linewidth]{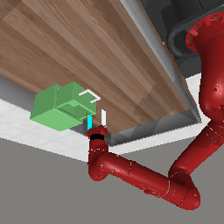} & \includegraphics[angle=180,width=0.158\linewidth]{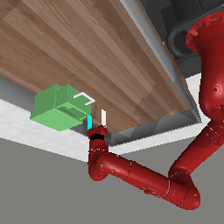} & \includegraphics[angle=180,width=0.158\linewidth]{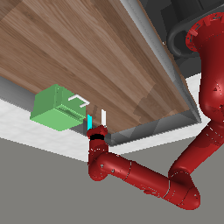} & \includegraphics[angle=180,width=0.158\linewidth]{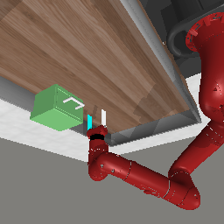}\\
\end{tabular}\par\medskip
\textbf{Faucet}: native joint coordinate (rad)\par\smallskip
\begin{tabular}{@{}cccccc@{}}
$0.000$ & $0.280$ & $0.560$ & $0.840$ & $1.120$ & $1.400$\\
\includegraphics[angle=180,width=0.158\linewidth]{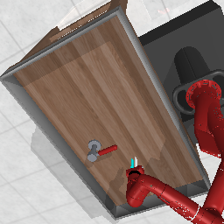} & \includegraphics[angle=180,width=0.158\linewidth]{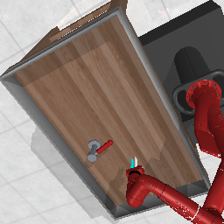} & \includegraphics[angle=180,width=0.158\linewidth]{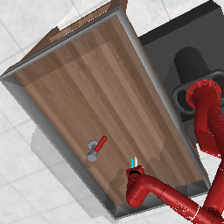} & \includegraphics[angle=180,width=0.158\linewidth]{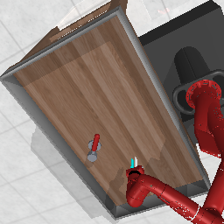} & \includegraphics[angle=180,width=0.158\linewidth]{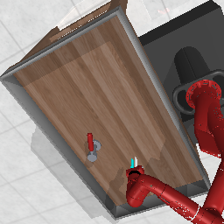} & \includegraphics[angle=180,width=0.158\linewidth]{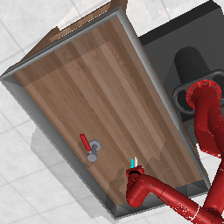}\\
\end{tabular}\par\medskip
\caption{Controlled observations at fixed robot contexts. Each row contains six coordinate settings from one development scene: Drawer seed 108300, robot-context frame 6; Faucet seed 110300, robot-context frame 12. Coordinates increase from left to right, follow the simulator's native sign convention, and are rounded for display. A fixed $180^\circ$ orientation correction presents the rendered images upright. These are illustrative examples; quantitative validation is reported in Appendices A and D.}
\label{fig:controlled-examples}
\end{figure}

Figure~\ref*{fig:controlled-examples} shows target-coordinate variation at a fixed robot context in one development scene per task. The selected groups are the first available scene and robot-context frame in numeric order for each task. Columns are ordered by the measured native joint coordinate. The supplementary package provides the 12 source images, full-precision coordinates, and their image hashes.

\end{document}